%% file: seg_arxiv.tex
\documentclass[10pt,twocolumn,letterpaper]{article}

\pdfoutput=1 

\usepackage{times}
\usepackage{epsfig}
\usepackage{graphicx}
\usepackage{amsmath}
\usepackage{amssymb}

\usepackage{color}

\usepackage[pagebackref=true,breaklinks=true,letterpaper=true,colorlinks,bookmarks=false]{hyperref}

\input{macros}

\begin{document}

\title{Co-segmentation for Space-Time Co-located Collections}
\author{Hadar Averbuch-Elor$^{1}$
        Johannes Kopf$^{2}$
        Tamir Hazan$^{3}$
        Daniel Cohen-Or$^{1}$ \vspace{3pt}
        \\ 
         $^1$Tel Aviv University \hspace{4pt}
         $^2$Facebook \hspace{4pt}
		 $^3$Technion -- Israel Institute Of Technology 
       }
\date{}
\maketitle

\begin{abstract}
\input{abstract}
\end{abstract}

\input{intro}

\input{related}
\input{iterations}

\input{background}

\input{single_template}

\input{prop_templates}

\input{evaluation}

\input{conclusions}


{\small
\bibliographystyle{ieee}
\bibliography{seg_siggraph}
}

\end{document}

%% file: macros.tex
\long\def\ignorethis#1{}

\usepackage{amsmath}
\usepackage{xfrac}
\usepackage{colortbl}

\usepackage{color}
\definecolor{gray}{rgb}{0.35,0.35,0.35}
\definecolor{lightgray}{rgb}{0,0,0}
\definecolor{blue}{rgb}{0,0,1}
\definecolor{white}{rgb}{1,1,1}

\usepackage{epsfig}
\usepackage{epstopdf}
\usepackage{subfig}
\usepackage{multirow}
\usepackage{array}
\usepackage{booktabs}
\newcommand{\ra}[1]{\renewcommand{\arraystretch}{#1}}

\newcommand{\whitetxt}[1]{{\color{white} #1}\normalfont}

\newbox\jsavebox
\newcommand{\jsubfig}[2]{%
	\sbox\jsavebox{#1}%
	\parbox[t]{\wd\jsavebox}{\centering\usebox\jsavebox\\#2}%
	}

%% file: abstract.tex
We present a co-segmentation technique for space-time co-located image collections. These prevalent collections capture various dynamic events, usually by multiple photographers, and may contain multiple co-occurring objects which are not necessarily part of the intended foreground object, resulting in ambiguities for traditional co-segmentation techniques. 
Thus, to disambiguate what the common foreground object is, we introduce a weakly-supervised technique, where we assume only a small seed, given in the form of a single segmented image.
We take a distributed approach, where local belief models are propagated and reinforced with similar images.  Our technique progressively expands the foreground and background belief models across the entire collection. The technique exploits the power of the entire set of image without building a global model, and thus successfully overcomes large variability in appearance of the common foreground object. We demonstrate that our method outperforms previous co-segmentation techniques on challenging space-time co-located collections, including dense benchmark datasets which were adapted for our novel problem setting.

%% file: intro.tex


\section{Introduction}

\begin{figure} 
\centering%
	\includegraphics[width=\columnwidth]{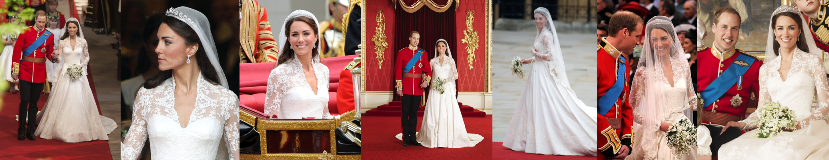}
	\includegraphics[width=\columnwidth]{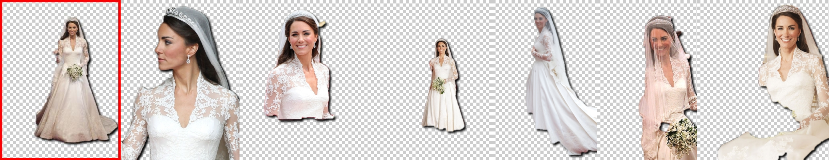}	
	\caption{The appearance of the Duchess of Cambridge varies throughout the images that capture her wedding ceremony. Starting from a single image template (marked with a red border), our method progressively expands the foreground belief model across the entire collection.} 
	\label{fig:teaser}
\end{figure}

Nowadays, \emph{Crowdcam} photography is both abundant and prevalent  \cite{basha2012photo,arpa2013crowdcam}. A crowd of people capturing various events form collections with great variety in content. However, they normally share a common theme. 
We refer to a collection of images that was captured about the same time and space as `'Space-time Co-located'' images, and we assume that such a co-located collection contains a significant subset of images that share a common foreground object, but other objects may also co-occur throughout the collection.  
See Figure \ref{fig:teaser} for such an example, where the Duchess of Cambridge is photographed in her wedding, and some of the images contain, for instance, her husband, Duke of Cambridge. 

Foreground extraction is one of the most fundamental problems in computer vision, receiving ongoing attention for several decades now. Technically, the problem of cutting out the common foreground object from a collection of images is known and has been referred to as co-segmentation \cite{faktor2013co,rubinstein2013unsupervised,cheng2011salient}. 
A traditional co-segmentation problem assumes that objects which are both co-occurring and salient necessarily belong to the foreground regions. However, the space-time co-location of the images leads to a more challenging setting, where the premise of common co-segmentation techniques is no longer valid, as the foreground object is not well-defined. 
Therefore, we ask the user to provide a segmented template image to specify what the intended foreground object is.

\begin{figure*}[t]
\centering
	\jsubfig{\includegraphics[height=4.0cm]{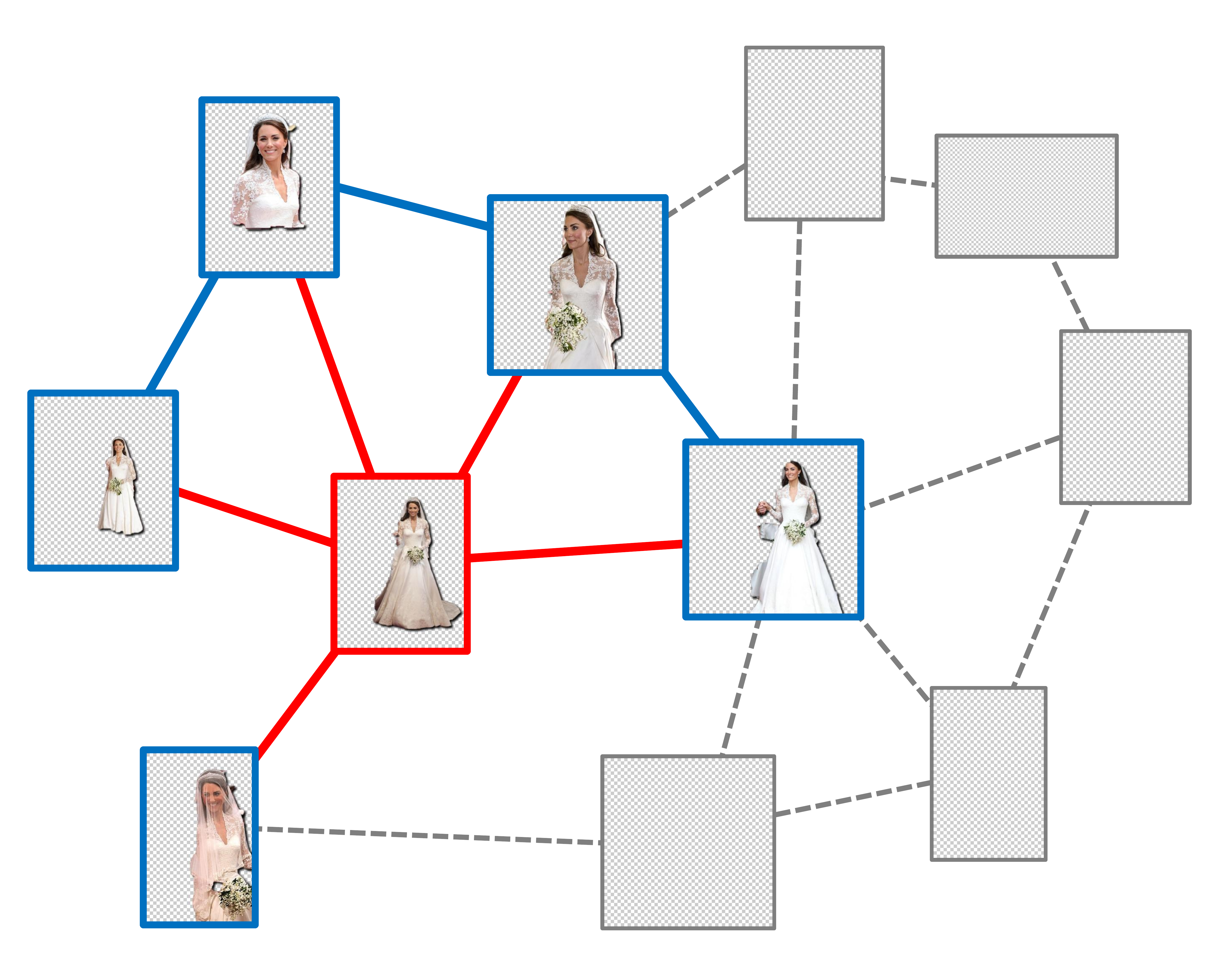}}
	{Iteration $i$}%
	\hspace{3pt}
	\jsubfig{\includegraphics[height=4.0cm]{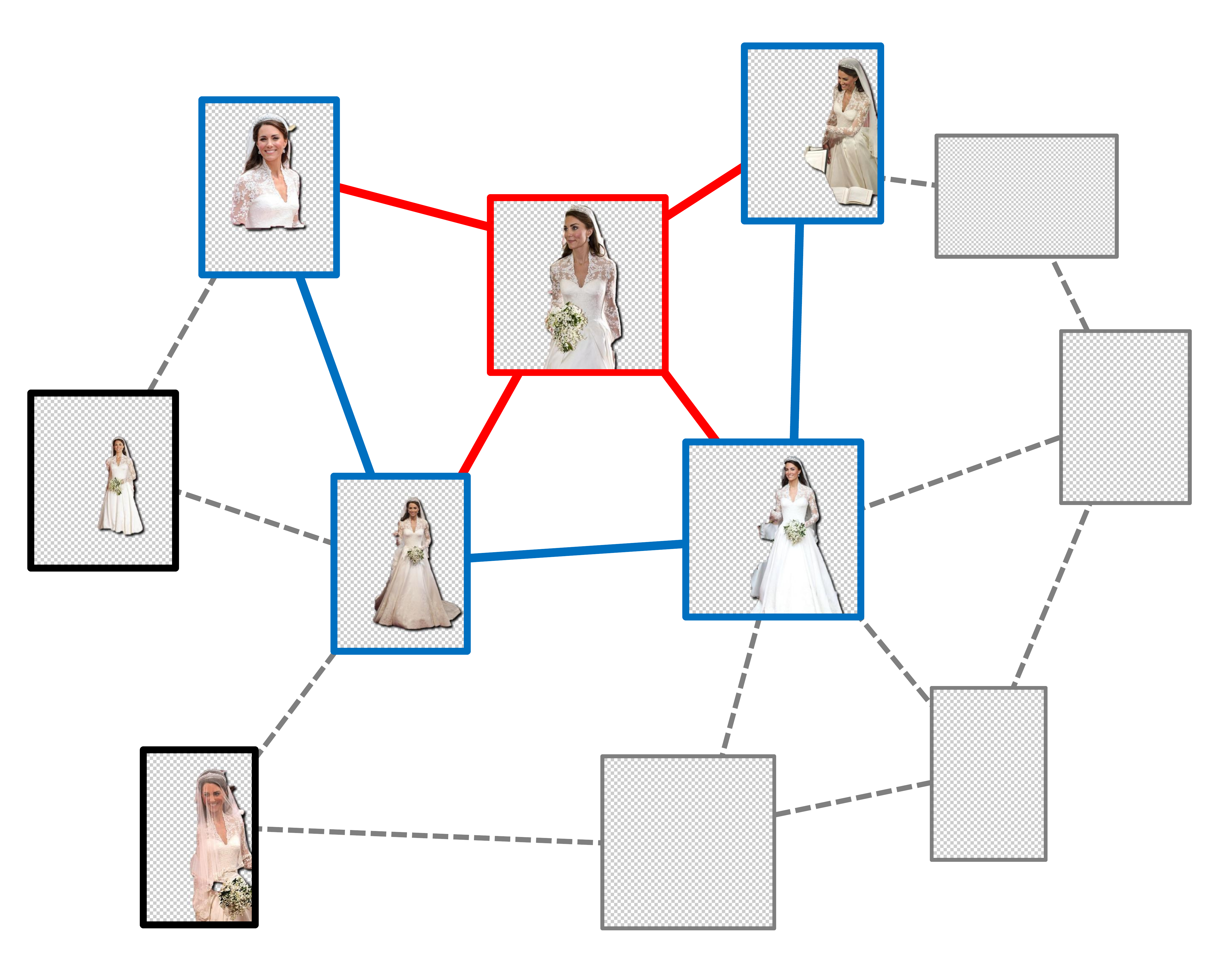}}
	{Iteration $i+1$}%
	\hspace{3pt}
	\jsubfig{\includegraphics[height=4.0cm]{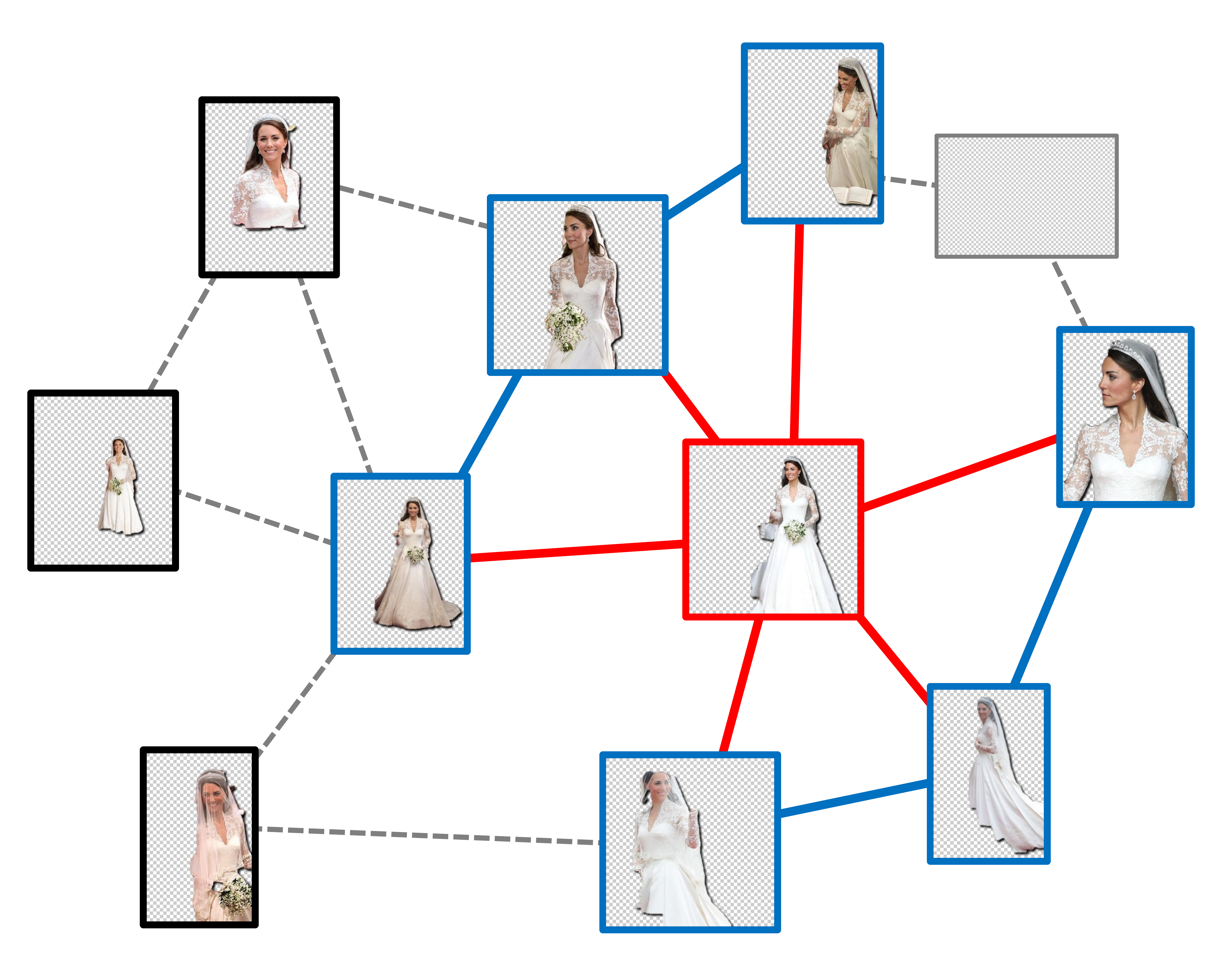}}
	{Iteration $i+2$}%
\vspace{-5pt}
	\caption{Our technique iteratively propagates beliefs to images (framed in blue) which are adjacent to the current seed (framed in red). In each iteration, object likelihood maps are first inferred from the seed image to each one of its adjacent images (illustrated with red edges) and then these maps are propagated across similar images (illustrated with blue edges) to reinforce the inference.  
}
\vspace{-16pt}
	\label{fig:image_graph1}
\end{figure*}

The foreground object varies considerably in appearance across the entire space-time co-located collection.
Thus, we do not use a single \emph{global} model to represent it, but instead take a distributed \emph{local} approach.
We decompose each image into parts at multiple scales. Parts store local beliefs about the foreground and background models.
These beliefs are iteratively propagated to similar parts within and among images.
In each iteration, one image is selected as the current seed. See Figure \ref{fig:image_graph1} which illustrates the progression of beliefs in the network of images. 

The propagation of beliefs from a given seed is formulated as a convex belief propagation (CBP) optimization. Foreground and background likelihood maps of neighboring images are first inferred independently (see Section \ref{sec:single}). These beliefs are then reinforced across images to consolidate local models and thus allow for more refined likelihood estimates (see Section \ref{sec:multiopt}). To allow for a joint image inference, we extend the 
CBP algorithm to include quadratic terms.

We show that when starting from a reliable seed model, we can progressively expand the foreground belief model across the entire collection. This gradual progression succeeds to co-segment the collection, outperforming state-of-the-art co-segmentation techniques on rich benchmark datasets which were adapted for our problem setting. We also provide an extensive evaluation on various space-time co-located collections which contain repeated elements that do not necessarily belong to the semantic foreground region. Our analysis demonstrates the advantages of our technique over previous methods, and in particular illustrates its robustness against significant cluttered backgrounds. 

The main contributions of our work are (i) the introduction of the novel co-segmentation problem for space-time co-located image collections, (ii) a distributed approach that can handle the great variability in appearance of the foreground object, and (iii) an extended variational scheme for propagating information within and across images.  


%% file: related.tex
\section{Related work}

Segmenting and extracting the foreground object from an image is a fundamental and challenging problem, which has received significant ongoing attention. Extracting the foreground object requires some guidance or supervision since in most cases it is unclear what the semantic intent is. When several images that share a common foreground are given, the problem is referred to as co-segmentation \cite{rother2006cosegmentation}. Many solutions have been proposed to the co-segmentation problem, which can be applied to image collections of varying sizes and characteristics
\cite{cheng2011salient,chang2011co,faktor2013co,rubinstein2013unsupervised,kim2011distributed}. Co-segmentation techniques learn the appearance commonalities in the collection to infer the common foreground object or objects. To initialize the learning process, unsupervised techniques are usually based on objectness \cite{vicente2011object} or visual saliency \cite{cheng2011salient,rubio2012unsupervised} cues to estimate the target object.    

State-of-the-art co-segmentation methods are based on recent advancements in feature matching and correspondence techniques \cite{rubio2012unsupervised,rubinstein2013unsupervised,faktor2013co}. 
Rubinstein \emph{et al}. \cite{rubinstein2013unsupervised} proposed to combine saliency and dense correspondences to co-segment large and noisy internet collections. Faktor and Irani \cite{faktor2013co} also use dense correspondences, however, they compute statistical significance of the shared regions, rather than computing saliency separately per image. These techniques are unsupervised, and they assume that recurrent and unique segments necessarily belong to the object of interest. However, in many collections this is not the case, and some minimal semantic supervision is then required. Batra et al. \cite{batra2010icoseg}, for example, aimed at \emph{topically related images} and their supervision was given in the form of multiple user scribbles. In our work, we deal with images that belong to the same instance, and not to a general class, which exhibit great variability in appearance. We use the segmentation of a single image in the collection to guide the process and target the intended object.

The work of Kim and Xing \cite{kim2012multiple} is most closely related to ours. In their work they address the problem of multiple foreground co-segmentation, where $K$ objects of interest repeatedly occur over an entire image collection. They show promising results when roughly $20\%$ percent of their images are manually annotated. In our work, we target a single object of interest, while space-time co-located collections often 
contain several repeated elements that clutter and distract common means to distinct the foreground. Unlike their global optimization framework, that solves for all the segmentations at once, our technique gradually progresses, and each image in turn guides the segmentation of its adjacent images. In this sense of progression, the method of Kuettel et. al \cite{kuettel2012segmentation} is similar to ours. However, their method is strongly based on the semantic hierarchy of ImageNet, while we aim at segmenting an unstructured space-time co-located image collection.

There are other space-time co-located settings where images share a common foreground. One setting is a video sequence \cite{ramakanth2014seamseg,fan2015jumpcut}, where the coherence among the frames is very high. It is worth noting that even then, the extraction of the foreground object is surprisingly difficult. Kim and Xing \cite{kim2013jointly} presented an approach to co-segment multiple photo streams, which are captured by various users at varying places and times. Similarly to our work, they also iteratively use a belief propagation model over the image graph. Another setting is multi-view object segmentation, e.g., \cite{gang2004silhouette,campbell2010automatic}, where the common object is captured from a calibrated set of cameras. These techniques commonly employ 3D reconstruction of the scene to co-segment the object of interest. In our setting, the images are rather sparse in scene-space and not necessarily captured all at once, which makes any attempt to reconstruct the target object highly improbable.



%% file: iterations.tex
\begin{figure}[t]
	\vspace{-5pt}
	\includegraphics[width=\columnwidth]{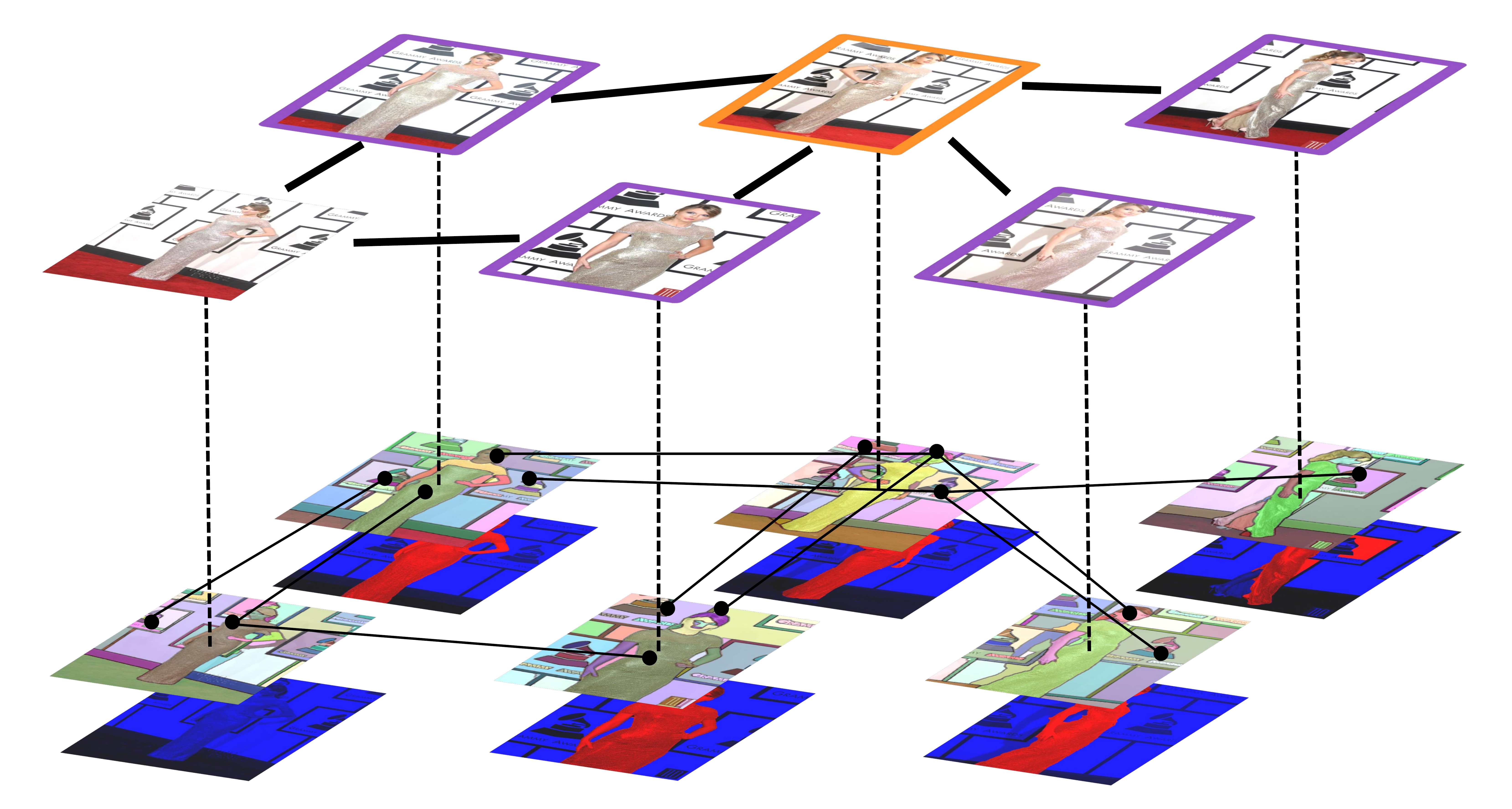}
	\vspace{-15pt}
	\caption{The image-level graph $G^i$ (on top) defines a topology of images over which local belief models are iteratively propagated. In each iteration, a seed image (marked with an orange border) propagates the F/B likelihood maps to its adjacent images (marked with a purple border). From these likelihood estimates, we extract the common foreground object (in red) and choose the next seed image.
}
	\label{fig:image_graph}
\end{figure}

\section{Co-segmentation using iterative propagation}
\label{sec:iterations}

We describe the foreground and background models, denoted by F and B, using local beliefs that are propagated within and across images.
To define a topology over which we propagate beliefs, we construct a parts-level graph  $G^p$, where nodes are image parts from all images, and edges connect corresponding parts in different images or spatially neighboring parts within an image. Furthermore, we define an associated image-level graph $G^i$, where the nodes correspond to the images, and two images are connected by an edge if there exists at least one edge in $G^p$ that connects the pair of images. The F/B likelihoods are iteratively propagated throughout the part-level graph $G^p$, while the propagation flow is determined according to the image-level graph $G^i$. The graph topology is illustrated in Figure \ref{fig:image_graph}.

In what follows, we first explicitly define the graph topology. We then describe how these beliefs are gradually spread across the entire image collection, starting from the user-segmented template image. 


\subsection{Propagation graph topology}
\label{sec:graph_topology}

The basis for the propagation are image parts.
To obtain the parts, we use the hierarchical image segmentation method of Arbel\'{a}ez \emph{et al}. \cite{APBMM2014}. We threshold the  ultrametric contour map, which defines the hierarchy of image regions, at a relatively fine level ($\lambda_i=0.15$). See Figure \ref{fig:pairwise_term} (on the left) for an illustration of the parts obtained at a number of different levels. The level we use for the image parts is illustrated in the left-most image. Although a fine level yields a large number and perhaps less meaningful parts, it should be noted that a coarser level often merges between foreground and background parts.

We construct a parts-level graph  $G^p$, where edges connect corresponding parts or spatially neighboring parts within an image. To compute reliable correspondences between image parts, we use the non-rigid dense correspondence technique (NRDC) \cite{hacohen2011non}, which outputs a confidence measure (with values between $0$ and $1$) along with each displacement value. We consider corresponding pixels to be those with a confidence which exceeds a certain threshold, which we set empirically to $0.5$. 

Two images are connected by an edge in the associated image-level graph $G^i$ if there exists at least one edge in $G^p$ that connects the pair of images.


\subsection{Iterative likelihood propagation}


We assign each part in $G^p$ a foreground likelihood. Initially all parts are equally likely to be foreground or background (except the parts in the user-segmented template image, whose F-likelihood is either exactly 0 or 1).

The likelihoods are iteratively propagated throughout the graphs.
In each iteration, a seed image is selected and its likelihoods are propagated to the adjacent neighbors in $G^i$.
In the first iteration, the seed image is always the user-segmented template image. In subsequent iterations the seed image randomly picked from the neighbors of the current seed. Within an iteration, the seed image likelihoods are considered fixed. Note that the template image likelihoods remain fixed throughout the whole algorithm.

The details of this propagation are described in the next section.
The new estimates are first derived separately, according to Section \ref{sec:single}, and are then jointly refined, according to Section \ref{sec:multiopt}. These new likelihood estimates are combined with previous estimates, where estimates are amortized along their propagation, and get exponentially lower weights over time, as we have more confidence in beliefs that are closer to our template image. 

After propagating the likelihoods, we update the foreground-background segmentation for the next seed using a modified implementation of graph-cuts \cite{kuettel2012segmentation}, where the unary terms are initialized according to the obtained likelihoods.  

The algorithm above is terminated once all images have been propagated to at least once. To avoid error accumulation, we execute the full pipeline multiple times (five in our implementation). The final results are obtained by averaging all the likelihood estimates follows by a graph-cut segmentation.

%% file: background.tex

\section{Likelihood inference propagation} 

Our algorithm uses convex belief propagation and further extends the variational approximate inference program to include quadratic terms. Therefore, in Section \ref{sec:background}, we briefly introduce notations used in later sections. In Section \ref{sec:single}, we present an approach to infer an object likelihood map of a single target image from a seed image. Finally, in Section \ref{sec:multiopt}, we introduce a technique to propagate the likelihood maps across similar images to improve the accuracy and reliability of these inferred maps.

\subsection{Convex belief propagation}
\label{sec:background}
%

Markov random fields (MRFs) consider joint distributions over discrete product spaces  $Y = Y_1 \times \cdots \times  Y_n$. The joint probability is defined by combining potential functions over subsets of variables. Throughout this work we consider two types of potential functions: single variable functions, $\theta_i(y_i)$, which correspond to the $n$ vertices in a graph, $i \in \{1,...,n\}$, and functions over pairs of variables $\theta_{i,j}(y_i,y_j)$ that correspond to the graph edges, $(i,j) \in E$. The joint distribution is then given by Gibbs probability model:
\begin{equation}
p(y_1,...,y_n) \propto \exp \Big(\sum_{i \in V} \theta_i(y_i) + \sum_{i,j \in E} \theta_{i,j} (y_i,y_j)\Big).
\end{equation}


Many computer vision tasks require to infer various quantities from the Gibbs distribution, e.g., the marginal probability $p(y_i) = \sum_{y \setminus y_i} p(y_1,...,y_n)$. 

Convex belief propagation \cite{Wainwright05-upper,Heskes06} is a message-passing algorithm that computes the optimal beliefs $b_i(y_i)$ which approximate the Gibbs marginal probabilities. Furthermore, under certain conditions, these beliefs are precisely the Gibbs marginal probabilities $p(y_i)$. For completeness, in the Supplementary Material, we define these conditions and explicitly describe the optimization program.

%% file: single_template.tex
\subsection{Single target image inference}
\label{sec:single}

\begin{figure} [t]
	\vspace{-5pt}
	\includegraphics[width=0.25\columnwidth]{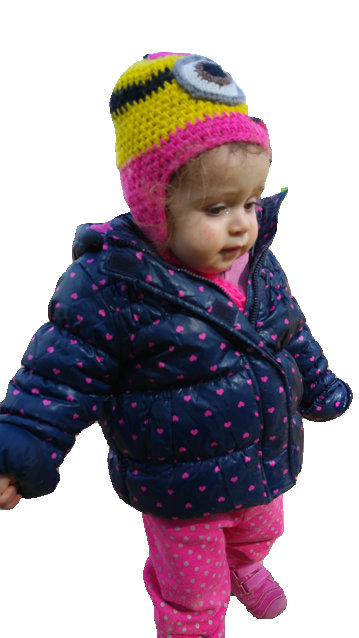}
	\includegraphics[width=0.74\columnwidth]{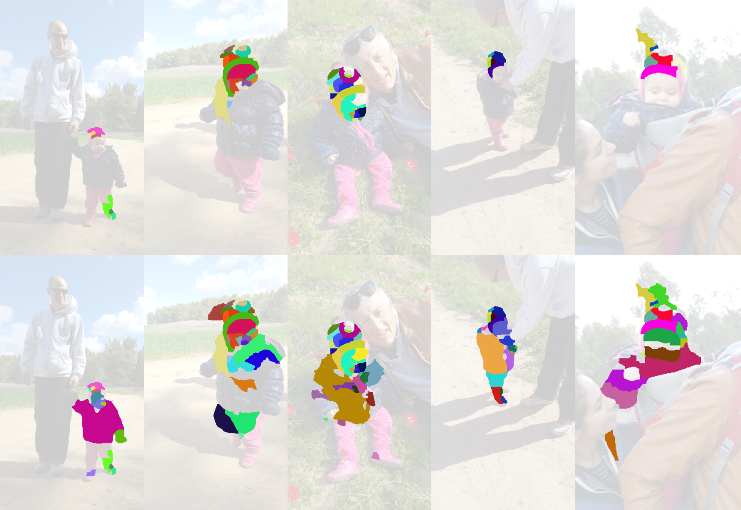}
	\vspace{-15pt}
	\caption{Corresponding foreground parts of adjacent images in $G^i$ according to $p_{corr}$ only (top row) vs our enriched compatibility measure (bottom row) that contains significantly more compatible parts. The foreground source is displayed on the left. 
\vspace{-8pt}	
	} 
	\label{fig:nrdc_corr}
\end{figure}

\begin{figure*} [t]
    \vspace{-5pt}
	\centering%
	\jsubfig{\includegraphics[height=4.2cm]{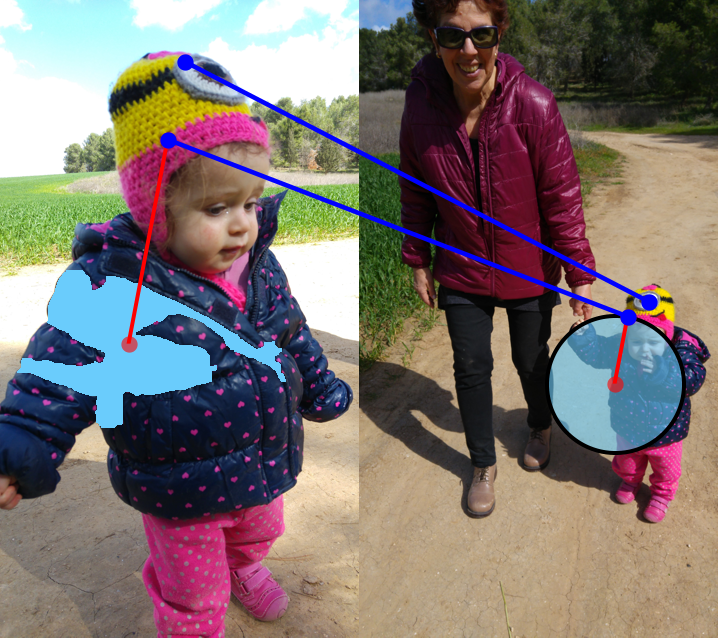}}
	{(a)}%
	\hfill%
	\jsubfig{
	\includegraphics[height=4.2cm]{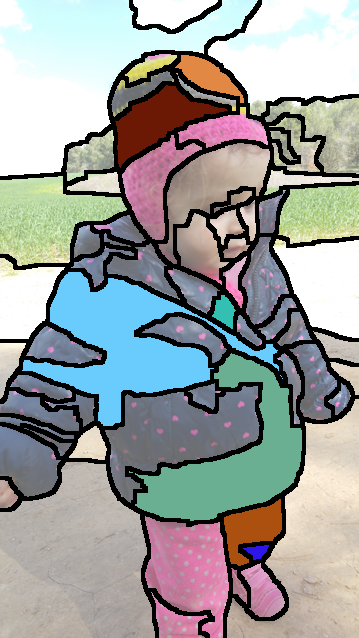}
	\includegraphics[height=4.2cm]{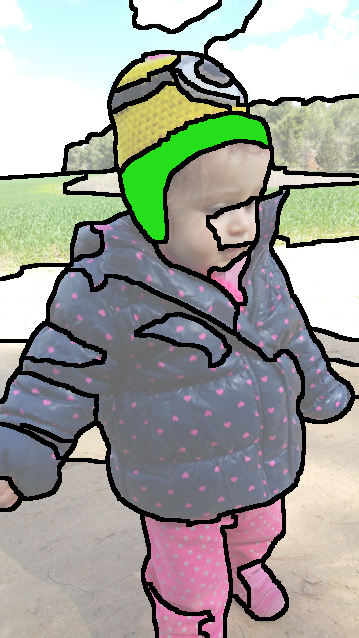}
	\includegraphics[height=4.2cm]{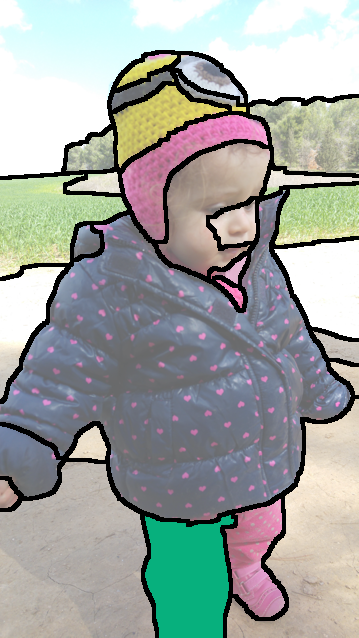}
	\includegraphics[height=4.2cm]{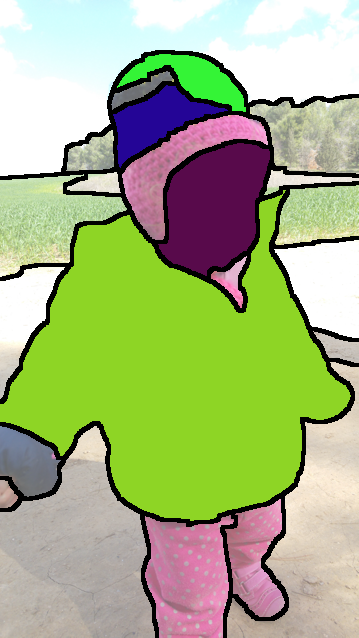}
	\includegraphics[height=4.2cm]{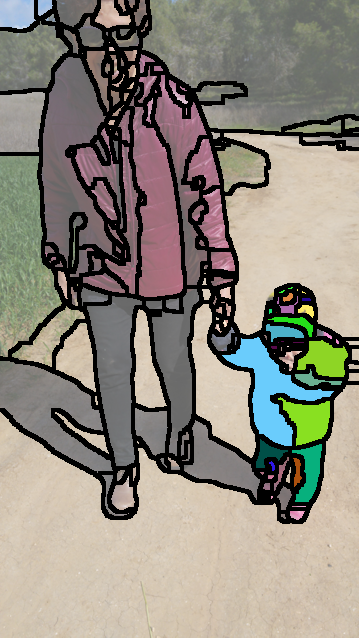}	
	}
	{(b)}%
	\vspace{-8pt}
	\caption{(a) Based on two reliable correspondences (in blue), the relative offset (in red) to a part (light blue) defines the region where the correspondent part is expected (marked with a light blue circle). 
	(b) The multi-scale parts of the source-seed image are matched to the parts of the target image (on the right). The parts that yield maximum compatibility are highlighted in unique colors (corresponding to the highlighted parts of the target image on the right). 
\vspace{-8pt}
}
	\label{fig:parts_unary}
\end{figure*}

In the following we present the basic component of our method, which infers an object likelihood map of a target image from an image seed. 
We construct a Markov random field (MRF) on the parts of the target image and use a convex belief propagation to infer the likelihood of these parts to be labeled as foreground.

Each part can be labeled as either foreground or background, i.e., $y_i \in \{-1,+1\}$. First, we describe the local potentials of each part $\theta_i(y_i)$, which describe the likelihood of the part to belong to the foreground or the background. 
Then, we describe the pairwise potentials $\theta_{i,j}(y_i,y_j)$ , which account for the spatial relations between adjacent parts.  We infer the foreground-background beliefs of the parts in the target image $b_i(y_i)$ by executing the standard convex belief propagation algorithm \cite{Wainwright05-upper,Heskes06}. 

\subsubsection{Local potentials}

\label{sec:local}

The local potentials $\theta_i(y_i)$ express the extent of agreement of a part with the foreground or background models. To define parts in the seed image, we use the technique of Arbel\'{a}ez et. al \cite{APBMM2014} at multiple levels to obtain a large bag of candidate parts of different scales. 
Let $i$ be a part in the target image, and $s$ be a part in the source image seed. Then for each source part $s$, we compute its compatibility with a target part $i$, and denote it by $p_{comp}(i,s)$. 

To construct the foreground/background likelihood of each part in the target image 
$i$, we consider the F/B parts of the source seed, and set
\begin{eqnarray*}
\theta_i(f) = \max_{s \in F}   p_{comp}(i,s) \hspace{3pt}  \textit{ and } \hspace{3pt}
\theta_i(b) = \max_{s \in B}   p_{comp}(i,s),
\end{eqnarray*}
where $f$ and $b$ are the two labels that can be assigned to $y_i$. 


We define our compatibility measure as follows:
\begin{equation}
p_{comp}(i,s) = p_{corr}(i,s)+\delta \cdot p_{sim}(i,s),
\end{equation}
where $\delta$ is a balancing coefficient that controls the amount of \emph{enrichment} of the available set of correspondences. 
The term $p_{corr}(i,s)$ measures the fraction of pixels that are matched between parts $i$ and $s$. This is measured according to 
\begin{equation}
p_{corr}(i,s) = N(i,s)~|s|^{-1},
\label{eq:pcorr}
\end{equation}
where $N(i,s)$ is the number of corresponding pixels, and $|s|$ is the number of pixels in part $s$. As mentioned before, the matching is based on NRDC. 

We identified that highly compatible parts are rather sparse, and thereby $p_{corr}(i,s)$ is almost always zero in many source-target pairs. Nonetheless, we can exploit these sparse correspondences to discover new compatible parts with the term $p_{sim}(i,s)$. See Figure \ref{fig:nrdc_corr} for an illustration of the compatible target parts in the foreground regions without (top row) and with (bottom row) our enrichment term $p_{sim}(i,s)$. In practice, since the background does not necessarily appear in both source and target image,  $\delta > 0$ only for regions $s \in F$.

In these foreground regions, the term $p_{sim}(i,s)$ aims at revealing a similarity between parts whose appearance and spatial displacement highly agree. Similarity in appearance, in our context, is measured according to the Bhattacharyya coefficient of the RGB histograms, following the method of \cite{ning2010interactive}. In order for parts $i$ and $s$ to highly agree in appearance, we further require that $i \in \mbox{top-k}(s)$, where the number of nearest neighbors is set to three. To recognize parts whose spatial displacement agree, we utilize the set of corresponding pixels in the foreground regions. We approximate the pixel values of the part corresponding to $s$ according to the known correspondences.  Formally, for each $s \in F$, let $i(s)$ be the estimated corresponding region in the target. Thus, for a similarity between parts $i$ and $s$, we require that $i \cap i(s) \ne \emptyset$.

To simplify computations, we assume $i(s)$ to be a circle within the target image, which we compute according to the closest and farthest foreground correspondences. These two corresponding points define a relative scale between the two images. To compute the circle center, we compute the relative offset from the closest corresponding point (using the relative scale). The radius is determined according to the distance to the nearest corresponding point in the target. See Figure \ref{fig:parts_unary}(a) for an illustration of the estimated corresponding region $i(s)$.

Putting it all together, $p_{sim}(i,s)$ is measured according to
{ \footnotesize
\[
p_{sim}(i,s)= \left\{
\begin{array}{cc}
\hspace{-7pt} \sum_{u=1}^{16^3}  \sqrt{hist(i)_u \cdot hist(s)_u} & \hspace{-3pt} i \in \mbox{top-k}(s), i\cap i(s) \ne \emptyset \\
 0 & \text{otherwise}.
\end{array}\right.
\]
}

In our experiments, we set $\delta = 0.1$ for all the foreground regions. See Figure \ref{fig:parts_unary}(b) for an illustration of the multi-scale source parts that obtained maximum compatibility with parts in the target image.

\subsubsection{Pairwise potentials}
\label{sec:pair}

\begin{figure*}
    \vspace{-5pt}
	\centering%
	\jsubfig{\includegraphics[height=3.25cm]{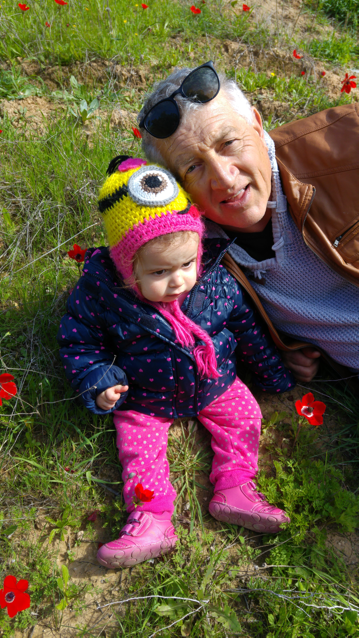}}
	{Target image}%
	\hfill%
	\jsubfig{
	\includegraphics[height=3.25cm]{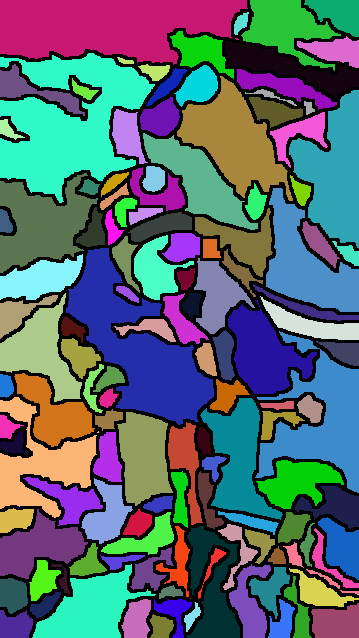}
	\includegraphics[height=3.25cm]{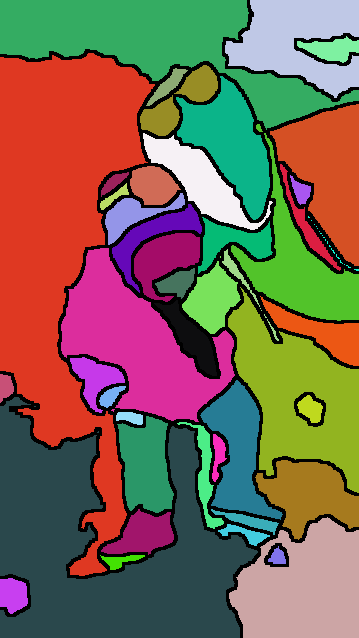}
	\includegraphics[height=3.25cm]{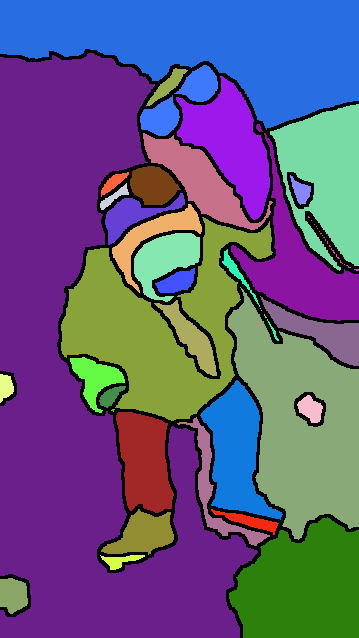}
	\includegraphics[height=3.25cm]{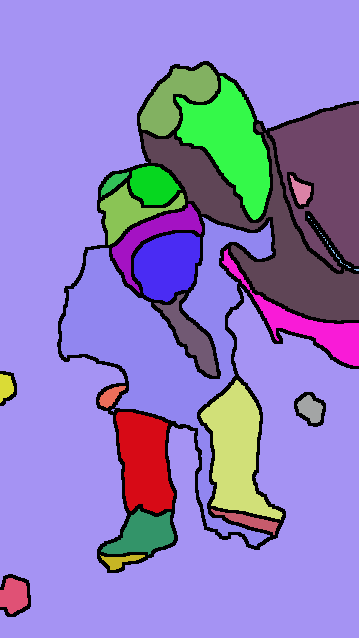}
	}
	{Target parts in different scales}%
	\hfill%
	\jsubfig{
	\includegraphics[height=3.25cm]{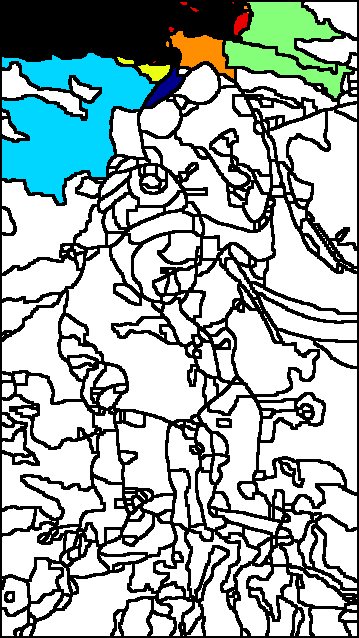}
	\includegraphics[height=3.25cm]{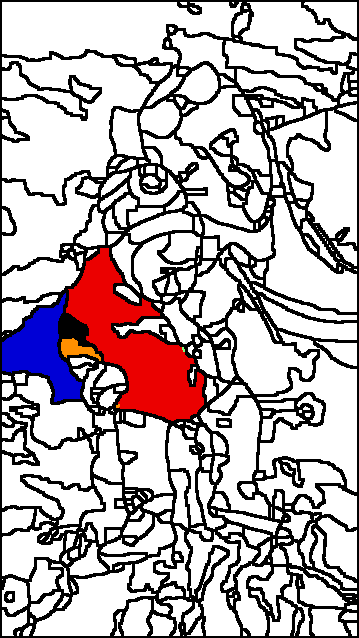}
	\includegraphics[height=3.25cm]{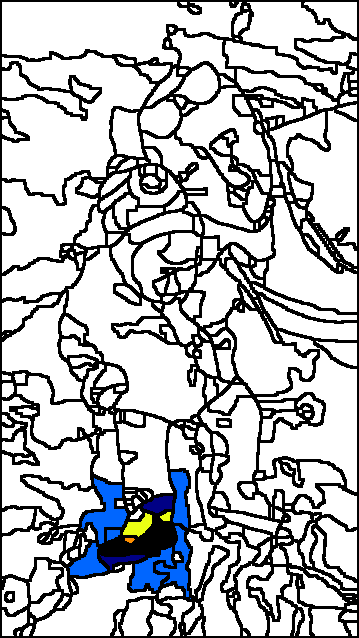}
	\includegraphics[height=3.25cm]{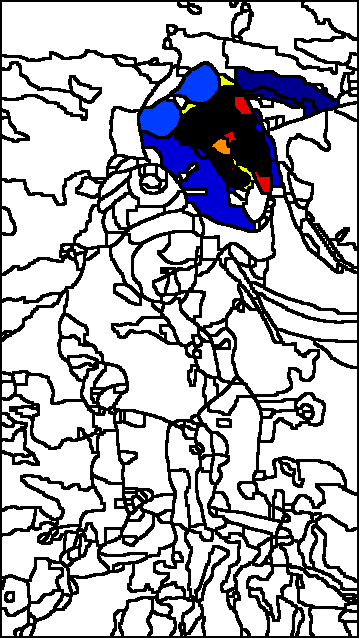}
	}
	{Visualizing the spatial consistencies}%
	\vspace{-6pt}
	\caption{Given the image on the left, parts are obtained using hierarchical segmentation. 
For each target image in the collection, we obtain a large number of parts, as in the left-most image. On the right are visualizations of various parts (colored in black) and the potentials which are induced to their neighboring parts. The neighboring parts are colored according to their proximities which are expressed in Equation \ref{eq:malik} (warm colors correspond to strong proximities while cool colors correspond to weaker proximities).  
}
	\label{fig:pairwise_term}
\end{figure*}

The pairwise potential function $\theta_{i,j}(y_i,y_j)$ induces spatial consistency from the part generation process within the image. As previously mentioned, we obtain parts at multiple scales by thresholding at varying levels $\lambda_i$ in the ultrametric contour map (see Figure \ref{fig:pairwise_term} for an illustration). 

To measure spatial consistencies between adjacent parts in the target, we can compute how \emph{quickly} these two parts merge into one by examining the level $\lambda_{merge}$ where the two parts become one. 
Hence, we define a pairwise relation between adjacent parts in each target image according to:
\begin{eqnarray}
\theta_{i,j}^{intra}(y_i,y_j) =  \exp \left( -\tau \left(\lambda_{merge}-\lambda _{min} \right) \right) \cdot  y_i y_j 
\label{eq:malik}
\end{eqnarray}
where the parameter $\tau=4$ was set empirically, and the finest level we examine to measure the spatial consistencies is $\lambda_{min}=0.2$ (a merge there would induce the strongest proximity between the parts). 
See the heat-maps in Figure \ref{fig:pairwise_term} for an illustration of these pairwise potentials on a few randomly-chosen target parts.
See Figure \ref{fig:pairwise_term} for an illustration of $\theta_{i,j}^{ucm}(y_i,y_j)$.

%% file: prop_templates.tex

\subsection{Joint multi-target inference}
\label{sec:multiopt}

In Section \ref{sec:single}, we presented our approach to infer an object likelihood map from a seed image. In our setting, similar regions may co-occur across multiple images. 
Therefore, to improve the accuracy and reliability of the likelihood maps obtained by a single inference step, we propagate the inferred maps onto adjacent images in the image graph $G^i$. The output beliefs of each inferred target image is sent to its neighbors as a heat-map (i.e., per part foreground-background probability). 
Thus, our likelihood maps are complemented with joint inference across images. 


To differentiate the different types of edges on $G^p$, we denote the edges that connect parts across images by $E^b$. 
A joint inference is encouraged by a pairwise potential function between matched parts in $E^b$. 
Since the labels satisfy $y_i,y_j \in \{-1,+1\}$, this can be done with the potential function
\begin{eqnarray}
\theta ^b _{i,j}(y_i,y_j) &=&  \Big(p_{corr}(i,j) + p_{corr}(j,i)\Big)  y_i y_j 
\label{eq:corr_parts}
\end{eqnarray}


Simply stated, the local potentials propagate the output beliefs of one target as input potentials of its neighboring images. 
Our intuition is that the output beliefs $b_i(y_i)$, which is concluded by running a convex belief propagation within its image, serves as a source seed signal to the neighboring image. This introduces novel non-linearities to the propagation algorithm. To control these non-linearities, we extend the variational approximate inference program 
to include quadratic terms. 

We also determine the conditions for which these quadratic terms still define a concave inference program and prove that repeated convex belief propagation iterations across images achieve its global optimum. For more details on our extended variational approximate inference program, together with an empirical evaluation of our inference technique,  please refer to the Supplementary Material, which can be found on the project website.

%
%
%
%

%% file: evaluation.tex
\section{Results and evaluation}
\label{sec:evaluation}

\begin{figure*} 
	\includegraphics[width=\textwidth]{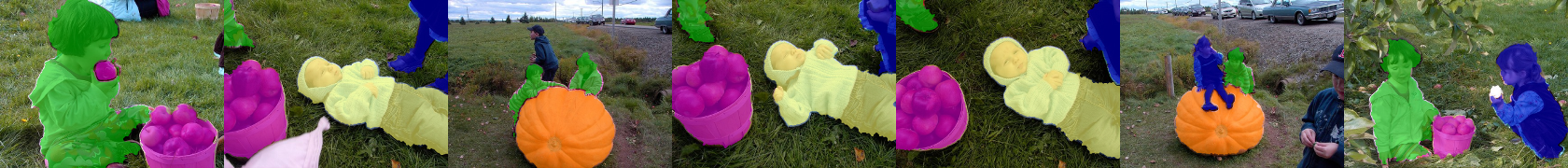}\\
	\includegraphics[width=\textwidth]{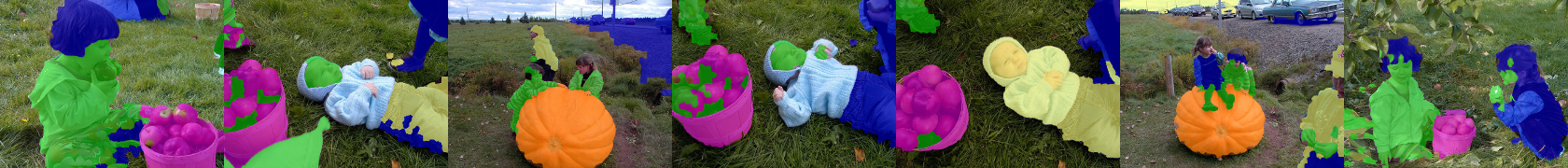}\\
	\includegraphics[width=\textwidth]{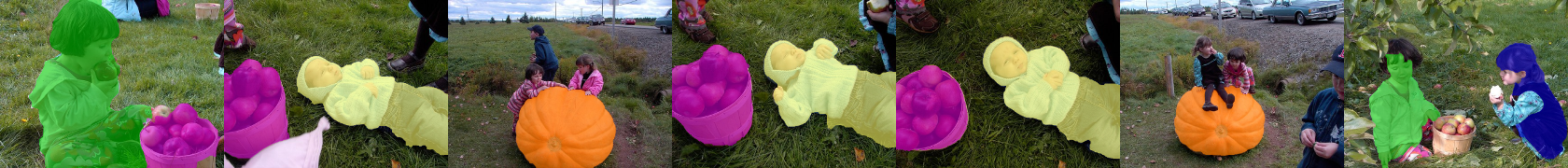}
   \vspace{-20pt}
   	\caption{Comparison to \cite{kim2012multiple} on their dataset. 
   	The top row illustrates the ground-truth labellings of multiple foreground objects: apple basket (in pink), pumpkin (in orange), baby (in yellow) and two girls (in green and blue). The bottom rows illustrate their results (middle row) and ours (bottom row).
On average, our method yields higher $P$ scores ($97.32\% \gg 93.11\%$) and comparable $J$ scores ($49.02\% \sim 49.61\%$).   
	\vspace{-10pt}
}
	\label{fig:mfc}
\end{figure*}

\begin{table}
\centering
\vspace{-5pt}
\ra{0.6}
\setlength{\tabcolsep}{2.3pt}
\resizebox{0.275\textwidth}{!}{%
\begin{tabular}{@{}lllrcrrrrrr@{}}
\toprule
\multicolumn{2}{c}{} &  & \multicolumn{2}{c}{\textsc{Bride}} & \phantom{}& \multicolumn{2}{c}{\textsc{Singer}} & \phantom{} & \multicolumn{2}{c}{\textsc{Toddler}}\\
 && & P & J &&  P & J &&  P & J \\ \midrule
 \cite{rubinstein2013unsupervised} && & 47.3 & 16.6 && 28.9 & 16.6  && 49.6 & 25.7\\ 
 \cite{faktor2013co} && & 71.1 & 42.9  && 68.4 & 30.4 && 81.8 & 44.1\\ 
 \cite{kim2012multiple} && & 63.9 & 27.4  && 88.8 & 63.4 && 66.7 & 38.9 \\ 
 Ours && & \textbf{88.9} & \textbf{76.3} &&\textbf{94.2} & \textbf{83.0}  && \textbf{94.5} & \textbf{74.2} \\

\bottomrule
\end{tabular}
}
\vspace{-6pt}
\footnotesize
\caption{Comparison against co-segmentation techniques on an annotated subset of our space-time co-located collections. \label{tab:comp}
\vspace{-15pt}
}
\end{table}

We empirically tested our algorithm on various datasets and compared our segmentation results to two state-of-the-art unsupervised co-segmentation techniques \cite{faktor2013co,rubinstein2013unsupervised} and to another semi-supervised co-segmentation technique \cite{kim2012multiple}. The merit of comparing our weakly-supervised technique with unsupervised ones is twofold; first, it serves as a qualitative calibration of the results on the new co-located setting, and second, it clearly demonstrates the necessity of some mininal supervision to define the semantic foreground region.

For all three methods we used the original implementations by the authors which are publicly available online. 
It should be noted that although \cite{kim2012multiple} discuss an unsupervised approach as well, they only provide implementations for the semi-supervised approach. To be compatible to our input, we provide their method with one input template mask. We measure the performance on different datasets, including benchmark datasets that were adapted for our novel problem setting. The full implementation of our method, along with the datasets that were used in the experiments, is available at our project website at: \url{https://cs.tau.ac.il/~averbuch1/coseg/}.

\vspace{-15pt}
\paragraph{Space-time images} We evaluated our technique on various challenging space-time co-located image collections depicting various dynamic events. Some of them (\textsc{Bride}, \textsc{Singer}, and \textsc{Broadway}) were downloaded from the internet, while others (\textsc{Toddler}, \textsc{Baby}, \textsc{Singer with Guitarist}, and \textsc{Peru}) were casually captured by multiple photographers.
These images contain repeated elements that do not necessarily belong to the semantic foreground region, and the appearance of the foreground varies greatly throughout the collections. We provide thumbnails for the \emph{full} seven sets, together with results and comparisons, on our project website. Please refer to these results for assessing the high quality of our results. 

\begin{figure*}[!t]
\vspace{-5pt}

\footnotesize
	\rotatebox[origin=c]{90}{\whitetxt{2222} \cite{rubinstein2013unsupervised}}   \hfill \includegraphics[width=0.97\textwidth]{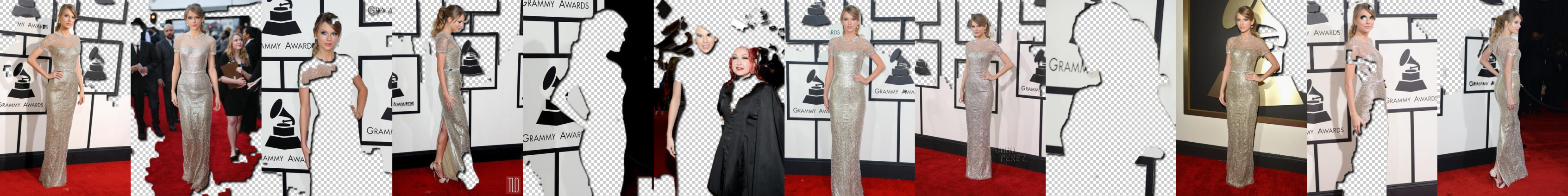} \\
	
	\vspace{-14pt}

	\rotatebox[origin=c]{90}{\whitetxt{2222} \cite{faktor2013co}}   \hfill \includegraphics[width=0.97\textwidth]{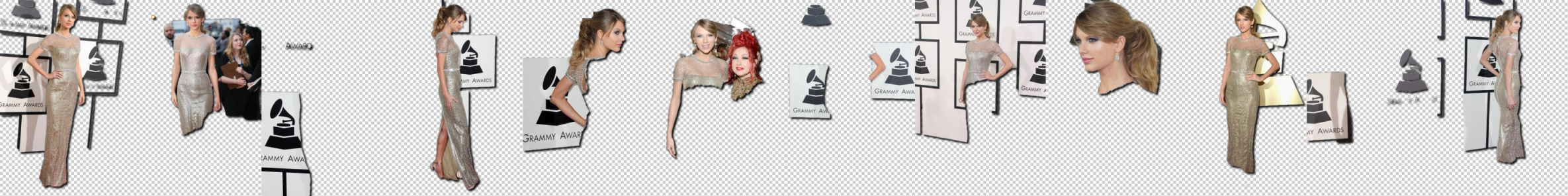} \\
	
	\vspace{-14pt}
	
	\rotatebox[origin=c]{90}{\whitetxt{2222} \cite{kim2012multiple}}   \hfill  \includegraphics[width=0.97\textwidth]{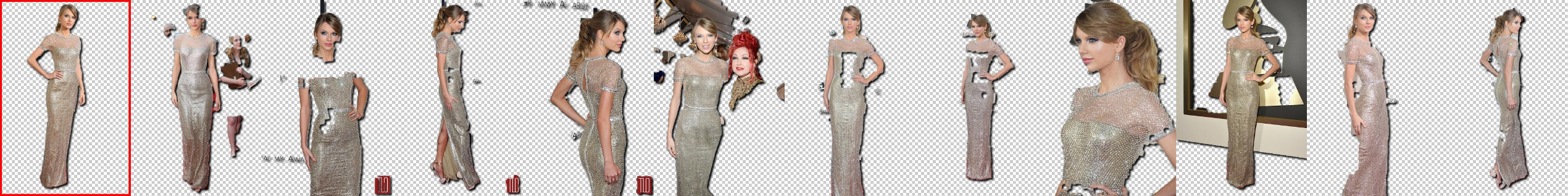}\\
	
	\vspace{-16pt}	
	
	\rotatebox[origin=c]{90}{\whitetxt{2222} Ours}   \hfill    \includegraphics[width=0.97\textwidth]{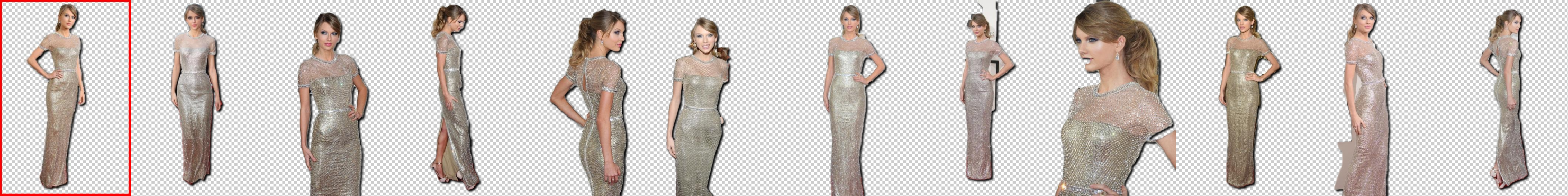} 
	
%
%
%
%
%
%
%
	
	\vspace{-13pt}
	
    	\rotatebox[origin=c]{90}{\whitetxt{2222} \cite{rubinstein2013unsupervised}}   \hfill \includegraphics[width=0.97\textwidth]{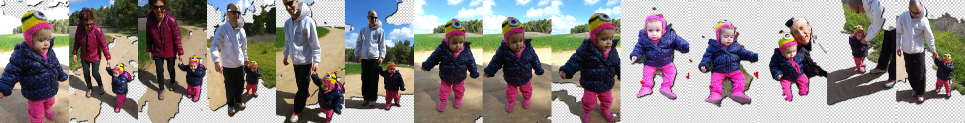} \\
    	
    \vspace{-14pt}
    	
	\rotatebox[origin=c]{90}{\whitetxt{2222} \cite{faktor2013co}} \hfill \includegraphics[width=0.97\textwidth]{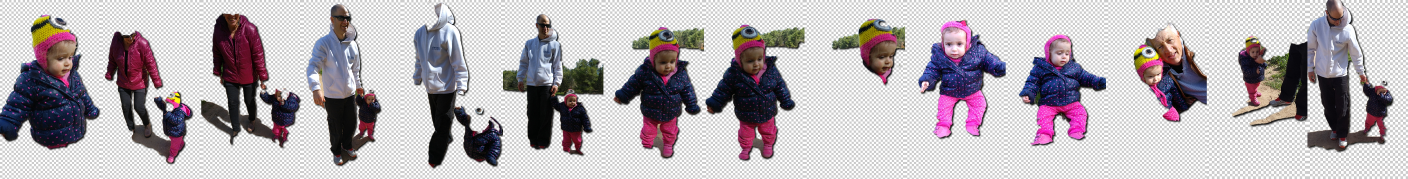} \\
	
	\vspace{-14pt}
	
	\rotatebox[origin=c]{90}{\whitetxt{2222} \cite{kim2012multiple}} \hfill \includegraphics[width=0.97\textwidth]{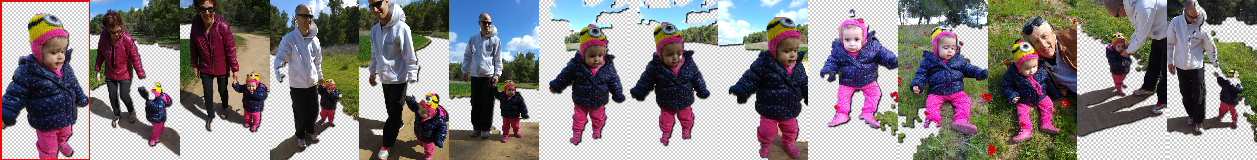}\\
	
	\vspace{-16pt}
	
	\rotatebox[origin=c]{90}{\whitetxt{2222} Ours} \hfill \includegraphics[width=0.97\textwidth]{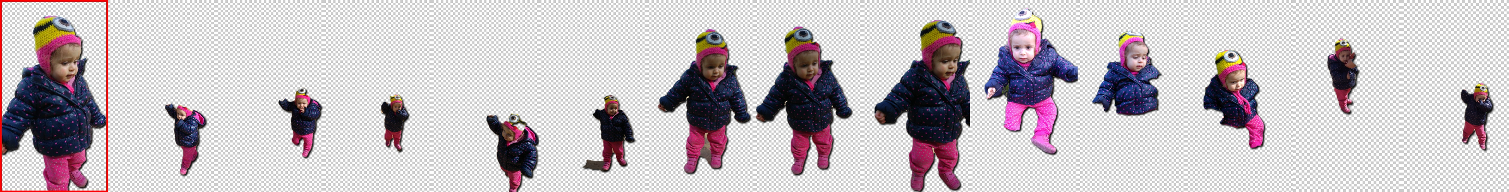} \\  

\vspace{-20pt}
	\caption{Qualitative comparison to state-of-the-art co-segmentation techniques on our space-time collections. 
	For both our technique and \cite{kim2012multiple}, the left-most image is provided as template. Please refer to our project website for an extensive and interactive comparison.
}
\vspace{-10pt}

	\label{fig:coseg}
\end{figure*}

For a quantitative analysis, we manually annotated the foreground regions of three of our collections (\textsc{Bride}, \textsc{Toddler}, and \textsc{Singer}), and report the precision $P$ (percentage of correctly labeled pixels) and Jaccard similarity $J$ (intersection over union of result and ground truth segmentations) as in previous works (see Table \ref{tab:comp}(b)). It should be noted that, to strengthen the evaluation, we perform three independent runs for the semi-supervised techniques, starting from different random seeds, and report the average scores. Figure \ref{fig:coseg} shows a sample of results, where the left-most image is provided as template for the semi-supervised techniques. As can be observed from our results, the unsupervised co-segmentation techniques fail almost completely on our co-located collections. Regarding the semi-supervised technique, as Figure \ref{fig:coseg} demonstrates, when both the foreground and background regions highly resemble those of their counterparts in the given template, then the results of \cite{kim2012multiple} are somewhat comparable to ours. As soon as the backgrounds differ or there are additional models that were not in the template, their method includes many outliers, as can be seen in Figure \ref{fig:coseg}. Unlike their method, we avoid defining strict global models that hold for all the images in the collection, and thus allow flexibility that is required to deal with the variability across the collection. 

\vspace{-15pt}
\paragraph{Multiple foreground objects}
We also compared our performance to \cite{kim2012multiple} using their data. We use their main example, which also corresponds to our problem setting. The results are displayed in Figure \ref{fig:mfc} where we mark the multiple foreground objects in different colors. We execute our method multiple times with different seeds to meet their input. 
As we can see here and in general, our method has less false-positives and is more resistant to cluttered backgrounds. If we are able to spread our beliefs towards the target image, then we succeed in capturing the object rather well. Quantitatively, our technique cuts the precision error by more than half (from $6.89\%$ down to $2.68\%$). However, if there is not enough confidence that reaches the target image, then the object remains undetected, as can be observed in the uncolored basket of apples in the rightmost image. 

\begin{figure*}
\begin{minipage}[t]{.729\textwidth} 
\vspace{0pt}
\includegraphics[width=\textwidth]{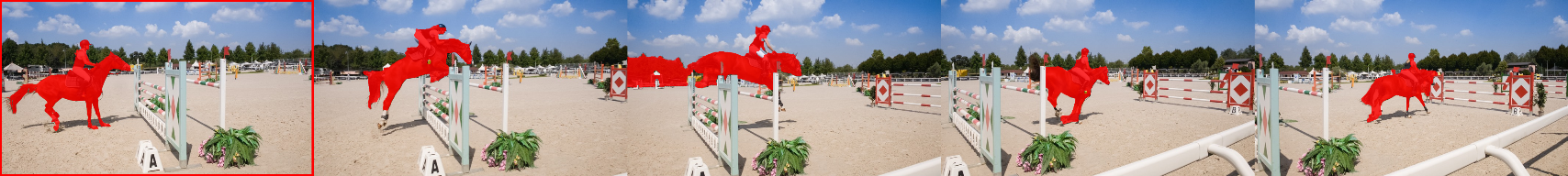}
\includegraphics[width=\textwidth]{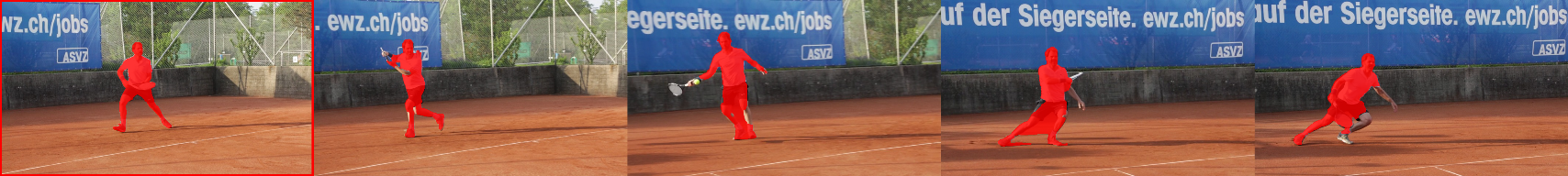}
\includegraphics[width=\textwidth]{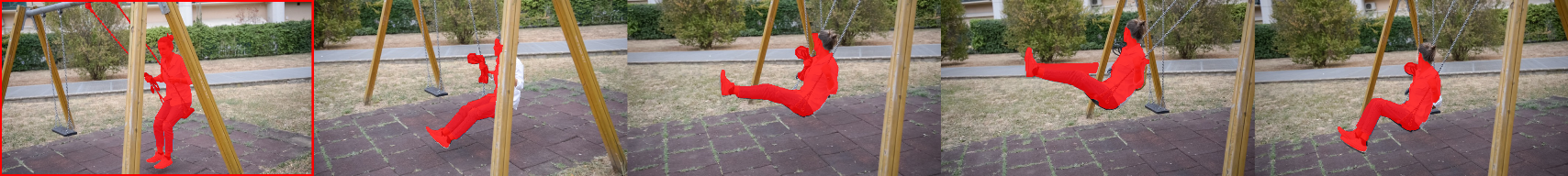}
\vspace{-18pt}
\caption{Qualitative results of our technique on a sequence of sparse frames sampled from the Davis dataset \cite{Perazzi2016}, where the first frame is provided as template. \label{fig:video} } 
\vspace{-5pt}
\captionof{table}{IoU scores on the sparsified DAVIS benchmarks. Following previous work, the scores are provided on a representative subset and the average computed over all 50 sequences.
\label{tab:video} }
\end{minipage}
\hspace{10pt}
\resizebox{0.215\textwidth}{!}{%
\begin{minipage}[t]{.27\textwidth} 
\centering
\hspace{-10pt}
\ra{0.6}
\setlength{\tabcolsep}{2.0pt}
\begin{tabular}{@{}llccccccccc@{}}
\toprule
 &&   \cite{rubinstein2013unsupervised} & \cite{faktor2013co} &  \cite{kim2012multiple} & Ours  \\ \midrule
 bear &&  0.19 & 0.05 & 0.73 & \textbf{0.92} \\
 blackswan && 0.30 & 0.07 & \textbf{0.74} & 0.64 \\
 bmx-trees && 0.04 & 0.06 & 0.08  & \textbf{0.27} \\
 bmx-bumps &&  0.03 & 0.17 & 0.24 & \textbf{0.28} \\
 breakdance-flare && 0.10 & 0.08 & \textbf{0.25} &0.08 \\
 breakdance &&  0.09 & 0.09 & \textbf{0.50} &0.29 \\
 bus &&  0.50 & \textbf{0.84} & 0.73 &\textbf{0.79}  \\
 dance-twirl &&  0.13 & 0.04 & 0.15 &\textbf{0.40}  \\
 libby && 0.38 & 0.14 & 0.29  &\textbf{0.41}\\
 dog &&  0.36 & 0.61 & \textbf{0.71} &0.57 \\
 drift-chicane &&  0.02 & 0.00 & \textbf{0.02} &0.00  \\
 drift-straight &&  0.13 & \textbf{0.31} & 0.11 & 0.26 \\
 mallard-water && 0.06 & 0.37 & 0.46  &\textbf{0.69}\\
 mallard-fly &&  0.01 & 0.05 & \textbf{0.47} &0.13  \\
 elephant &&  0.13 & 0.00 & 0.28 &\textbf{0.45}  \\
 flamingo &&  0.18 & 0.23 & 0.43 & \textbf{0.64} \\
 goat &&  0.07 & 0.04 & 0.42 &\textbf{0.64}  \\
 hike &&  0.17 & 0.00 & 0.36 & \textbf{0.89}  \\
 paragliding &&  0.30 & 0.13 & 0.80  & \textbf{0.82}\\
 soccerball && 0.02 & 0.00 & 0.37 & \textbf{0.67} \\
 surf &&  0.12 & 0.94 & 0.63 &\textbf{0.96} \\
 \midrule
 Average &&  0.16 & 0.22 & 0.36 & \textbf{0.53} \\

\bottomrule
\end{tabular}
\end{minipage}
}
\end{figure*}

\vspace{-15pt}
\paragraph{Sampled video collections}
The DAVIS dataset \cite{Perazzi2016} is a recent benchmark for video segmentation techniques, containing 50 sequences that exhibit various challenges including occlusions and appearance changes. The dataset comes with per-frame, per-pixel ground-truth annotations. We \emph{sparsified} these sequences (taking every 10$^{th}$ frame) to construct a large number of datasets that are somewhat related to our problem setting. Table \ref{tab:video} shows the intersection-over-union (IoU) scores on a representative subset and the average over all 50 collections. 
Similar to the input provided to video segmentation techniques in the mask propagation task, we also provide the semi-supervised techniques with a manual segmentation of the first frame. However, on our sparsified collections, subsequent frames are quite different, as illustrated in Figure \ref{fig:video}. 

Our extensive evaluation on the adapted DAVIS benchmark clearly illustrates, first of all, the difficulty of the problem setting, as the image structure is not temporally-coherent, and unlike dense video techniques, we cannot benefit from any temporal priors. Furthermore, it demonstrates the robustness of our technique, as it achieves the highest scores on most of the datasets, as well as the highest average score on all 50 collections.


%% file: conclusions.tex

\section{Conclusions and future work}
\label{sec:conclusions}

In this work, we have presented a co-segmentation method that takes a distributed approach. Common co-segmentation methods gather information from all the image in the collection, analyze it globally, building a common model, and then infer the common foreground objects in all, or part of, the images. Here, there is no global model. The beliefs are propagated across the collection without forming a global model of the foreground object. Each image independently, collects the beliefs from its neighbors, and consequentially infers its own model for the foreground object. Although our method is distributed, currently there is a seed model, which clearly does not concur to the claim of having a distributed method. However, some supervision is necessarily required to define the semantic target model. Currently, it is provided as a single segmented image, but the seed model can possibly be provided in other forms.

We have shown that our approach outperforms state-of-the-art co-segmentation methods. However, as our results demonstrate, there are limitations as the object cut-outs are imperfect. The entire object is not always inferred and also portions of the background may contaminate the extracted object. To alleviate these limitations, there are two possible avenues for future research: (i) one in high level, to better learn the semantics of the object, perhaps using data-driven approaches, e.g., convolutional networks, and (ii) in low level, seeking for better alternatives to graph-cuts and its inherent limitations.  


In the future, we hope to explore our approach on massive collections, which may include thousands of photographs capturing interesting dynamic events. For example, a collection of images of a parade, where a 3D reconstruction is not applicable. The larger number of images is not just a quantitative difference, but qualitative as well, as the collection can become dense with stronger local connections. For such massive collections, the foreground object does not have to be only a single object. We can propagate multi-target beliefs over the image network, like we demonstrated in our comparison to Kim and Xing \cite{kim2012multiple}. Finally, the distributed nature of our method, leads itself to parallel computation, which can be effective for large scale collections.


%% file: seg_arxiv.bbl
\begin{thebibliography}{10}\itemsep=-1pt

\bibitem{arpa2013crowdcam}
A.~Arpa, L.~Ballan, R.~Sukthankar, G.~Taubin, M.~Pollefeys, and R.~Raskar.
\newblock Crowdcam: Instantaneous navigation of crowd images using angled
  graph.
\newblock In {\em 3D Vision-3DV 2013, 2013 International Conference on}, pages
  422--429. IEEE, 2013.

\bibitem{basha2012photo}
T.~Basha, Y.~Moses, and S.~Avidan.
\newblock Photo sequencing.
\newblock In {\em Computer Vision--ECCV 2012}, pages 654--667. Springer, 2012.

\bibitem{batra2010icoseg}
D.~Batra, A.~Kowdle, D.~Parikh, J.~Luo, and T.~Chen.
\newblock icoseg: Interactive co-segmentation with intelligent scribble
  guidance.
\newblock In {\em Computer Vision and Pattern Recognition (CVPR), 2010 IEEE
  Conference on}, pages 3169--3176. IEEE, 2010.

\bibitem{campbell2010automatic}
N.~D. Campbell, G.~Vogiatzis, C.~Hern{\'a}ndez, and R.~Cipolla.
\newblock Automatic 3d object segmentation in multiple views using volumetric
  graph-cuts.
\newblock {\em Image and Vision Computing}, 28(1):14--25, 2010.

\bibitem{chang2011co}
K.-Y. Chang, T.-L. Liu, and S.-H. Lai.
\newblock From co-saliency to co-segmentation: An efficient and fully
  unsupervised energy minimization model.
\newblock In {\em Computer Vision and Pattern Recognition (CVPR), 2011 IEEE
  Conference on}, pages 2129--2136. IEEE, 2011.

\bibitem{cheng2011salient}
M.-M. Cheng, N.~J. Mitra, X.~Huang, P.~H. Torr, and S.-M. Hu.
\newblock Global contrast based salient region detection.
\newblock {\em IEEE Transactions on Pattern Analysis and Machine Intelligence},
  37(3):569--582, 2015.

\bibitem{faktor2013co}
A.~Faktor and M.~Irani.
\newblock Co-segmentation by composition.
\newblock In {\em Proceedings of the IEEE International Conference on Computer
  Vision}, pages 1297--1304, 2013.

\bibitem{fan2015jumpcut}
Q.~Fan, F.~Zhong, D.~Lischinski, D.~Cohen-Or, and B.~Chen.
\newblock Jumpcut: non-successive mask transfer and interpolation for video
  cutout.
\newblock {\em ACM Transactions on Graphics (TOG)}, 34(6):195, 2015.

\bibitem{gang2004silhouette}
Z.~Gang and Q.~Long.
\newblock Silhouette extraction from multiple images of an unknown background.
\newblock In {\em Proceedings of the Asian Conference of Computer Vision}.
  Citeseer, 2004.

\bibitem{hacohen2011non}
Y.~HaCohen, E.~Shechtman, D.~B. Goldman, and D.~Lischinski.
\newblock Non-rigid dense correspondence with applications for image
  enhancement.
\newblock {\em ACM transactions on graphics (TOG)}, 30(4):70, 2011.

\bibitem{Heskes06}
T.~Heskes.
\newblock {Convexity arguments for efficient minimization of the Bethe and
  Kikuchi free energies}.
\newblock {\em Journal of Artificial Intelligence Research}, 26(1):153--190,
  2006.

\bibitem{kim2012multiple}
G.~Kim and E.~P. Xing.
\newblock On multiple foreground cosegmentation.
\newblock In {\em Computer Vision and Pattern Recognition (CVPR), 2012 IEEE
  Conference on}, pages 837--844. IEEE, 2012.

\bibitem{kim2013jointly}
G.~Kim and E.~P. Xing.
\newblock Jointly aligning and segmenting multiple web photo streams for the
  inference of collective photo storylines.
\newblock In {\em Proceedings of the IEEE Conference on Computer Vision and
  Pattern Recognition}, pages 620--627, 2013.

\bibitem{kim2011distributed}
G.~Kim, E.~P. Xing, L.~Fei-Fei, and T.~Kanade.
\newblock Distributed cosegmentation via submodular optimization on anisotropic
  diffusion.
\newblock In {\em Computer Vision (ICCV), 2011 IEEE International Conference
  on}, pages 169--176. IEEE, 2011.

\bibitem{kuettel2012segmentation}
D.~Kuettel, M.~Guillaumin, and V.~Ferrari.
\newblock Segmentation propagation in imagenet.
\newblock In {\em Computer Vision--ECCV 2012}, pages 459--473. Springer, 2012.

\bibitem{ning2010interactive}
J.~Ning, L.~Zhang, D.~Zhang, and C.~Wu.
\newblock Interactive image segmentation by maximal similarity based region
  merging.
\newblock {\em Pattern Recognition}, 43(2):445--456, 2010.

\bibitem{Perazzi2016}
F.~Perazzi, J.~Pont-Tuset, B.~McWilliams, L.~V. Gool, M.~Gross, and
  A.~Sorkine-Hornung.
\newblock A benchmark dataset and evaluation methodology for video object
  segmentation.
\newblock In {\em Computer Vision and Pattern Recognition}, 2016.

\bibitem{APBMM2014}
J.~Pont-Tuset, P.~Arbelaez, J.~T. Barron, F.~Marques, and J.~Malik.
\newblock Multiscale combinatorial grouping for image segmentation and object
  proposal generation.
\newblock {\em IEEE transactions on pattern analysis and machine intelligence},
  39(1):128--140, 2017.

\bibitem{ramakanth2014seamseg}
S.~A. Ramakanth and R.~V. Babu.
\newblock Seamseg: Video object segmentation using patch seams.
\newblock In {\em Computer Vision and Pattern Recognition (CVPR), 2014 IEEE
  Conference on}, pages 376--383. IEEE, 2014.

\bibitem{rother2006cosegmentation}
C.~Rother, T.~Minka, A.~Blake, and V.~Kolmogorov.
\newblock Cosegmentation of image pairs by histogram matching-incorporating a
  global constraint into mrfs.
\newblock In {\em Computer Vision and Pattern Recognition, 2006 IEEE Computer
  Society Conference on}, volume~1, pages 993--1000. IEEE, 2006.

\bibitem{rubinstein2013unsupervised}
M.~Rubinstein, A.~Joulin, J.~Kopf, and C.~Liu.
\newblock Unsupervised joint object discovery and segmentation in internet
  images.
\newblock In {\em Computer Vision and Pattern Recognition (CVPR), 2013 IEEE
  Conference on}, pages 1939--1946. IEEE, 2013.

\bibitem{rubio2012unsupervised}
J.~C. Rubio, J.~Serrat, A.~L{\'o}pez, and N.~Paragios.
\newblock Unsupervised co-segmentation through region matching.
\newblock In {\em Computer Vision and Pattern Recognition (CVPR), 2012 IEEE
  Conference on}, pages 749--756. IEEE, 2012.

\bibitem{vicente2011object}
S.~Vicente, C.~Rother, and V.~Kolmogorov.
\newblock Object cosegmentation.
\newblock In {\em Computer Vision and Pattern Recognition (CVPR), 2011 IEEE
  Conference on}, pages 2217--2224. IEEE, 2011.

\bibitem{Wainwright05-upper}
M.~J. Wainwright, T.~S. Jaakkola, and A.~S. Willsky.
\newblock {A new class of upper bounds on the log partition function}.
\newblock {\em Trans. on Information Theory}, 51(7):2313--2335, 2005.

\end{thebibliography}
